\documentclass{article}

\usepackage{microtype}
\usepackage{graphicx}
\usepackage{booktabs} 
\usepackage{subcaption}
\usepackage{bm}

\usepackage{hyperref}
\usepackage{amsfonts}

 \usepackage[accepted]{icml2024}

\usepackage{amsmath}
\usepackage{amssymb}
\usepackage{mathtools}
\usepackage{amsthm}

\newcommand{\R}{\mathbb{R}}

\usepackage[capitalize,noabbrev]{cleveref}

\usepackage{acro}

\DeclareAcronym{urbf}{
    short = U-RBF,
    long = Univariate Radial Basis Function,
}

\DeclareAcronym{mrbf}{
    short = M-RBF,
    long = Multivariate Radial Basis Function,
}

\DeclareAcronym{rbf}{
    short = RBF,
    long = Radial Basis Function,
}

\DeclareAcronym{ffn}{
    short = FFM,
    long = Fourier Feature Mapping Network,
}

\DeclareAcronym{uffn}{
    short = U-FFN,
    long = Univariate Fourier Feature Mapping Network,
}

\DeclareAcronym{mlp}{
    short = MLP,
    long = Multi-Layer Perceptron,
}

\DeclareAcronym{pmlb}{
    short = PMLB,
    long = Penn Machine Learning Benchmarks,
}

\DeclareAcronym{svr}{
    short = SVR,
    long = Support Vector Regression,
}

\theoremstyle{plain}
\newtheorem{theorem}{Theorem}[section]

\newtheorem{corollary}[theorem]{Corollary}
\theoremstyle{definition}
\newtheorem{definition}[theorem]{Definition}

\theoremstyle{remark}

\usepackage[textsize=tiny]{todonotes}

\icmltitlerunning{Univariate Radial Basis Function Layers}

\begin{document}

\twocolumn[
\icmltitle{Univariate Radial Basis Function Layers: \\Brain-inspired Deep Neural Layers for Low-Dimensional Inputs}



\icmlsetsymbol{equal}{*}

\begin{icmlauthorlist}
\icmlauthor{Daniel Jost}{equal,yyy}
\icmlauthor{Basavasagar Patil}{equal,yyy}
\icmlauthor{Chris Reinke}{yyy}
\icmlauthor{Xavier Alameda-Pineda}{yyy}
\end{icmlauthorlist}

\icmlaffiliation{yyy}{INRIA Grenoble, LJK, UGA}

\icmlcorrespondingauthor{Daniel Jost}{jost.daniel@protonmail.com}
\icmlcorrespondingauthor{Basavasagar Patil}{agarbasava8@gmail.com}

\icmlkeywords{Machine Learning, ICML}

\vskip 0.3in
]



\printAffiliationsAndNotice{\icmlEqualContribution} 

\begin{abstract}
Deep Neural Networks (DNNs) became the standard tool for function approximation with most of the introduced architectures being developed for high-dimensional input data. However, many real-world problems have low-dimensional inputs for which the standard \ac{mlp}
are a common choice. An investigation into specialized architectures is missing.
We propose a novel DNN input layer called \ac{urbf} Layer as an alternative.
Similar to sensory neurons in the brain, the U-RBF Layer processes each individual input dimension with a population of neurons whose activations depend on different preferred input values.
We verify its effectiveness compared to MLPs and other state-of-the-art methods in low-dimensional function regression tasks.
The results show that the \ac{urbf} Layer is especially advantageous when the target function is low-dimensional and high-frequent. The code can be found at: \url{https://bit.ly/3vXyfRl}
\end{abstract}

\section{Introduction}





 

Neural networks became the fundamental tool for function approximation solving a wide range of problems, from natural language processing~\citep{openai2023gpt4} to computer vision~\citep{zhai2022scaling} and control~\citep{schrittwieser2020mastering}. 
Although a \ac{mlp} with a single large hidden layer can theoretically approximate any function~\citep{Hornik1989MultilayerFN}, training such simple models even with several layers is often not successful. 
Different architectures have been investigated allowing the training of networks often depending on their input data type.
Existing approaches were mainly developed for high-dimensional inputs such as Convolutional Neural Networks (CNNs) for images~\citep{lecun1989backpropagation,ciresan2011flexible} or Transformers for sequential data such as sentences~\citep{vaswani2017attention}. 

However, many important datasets and tasks are not high-dimensional.
The \ac{pmlb} \cite{romano_pmlb_2021} introduces a diverse set of real-world classification and regression problems with many having $15$ or fewer features and therefore being low-dimensional.
To be more specific, financial datasets for classification or regression such as fraud detection~\citep{al2021financial} usually have low-dimensional inputs.
For example, the credit approval dataset for classification of the UCI Machine Learning Repository \citep{Asuncion2007UCIML} which is also part of the \ac{pmlb}, consists only of 15 attributes (continuous and binary).

In recent advancements, the ability to learn low-dimensional target functions has been improved using \ac{ffn} \citep{tancik2020fourier}. This approach specifically focuses on the 'spectral bias' issue, commonly encountered in conventional \ac{mlp}, which refers to the insufficient ability to approximate high-frequency functions. The idea is to leverage Fourier features in a predefined frequency range and pass them to a standard \ac{mlp}. The selection of the frequency range, crucial for optimal performance, is not straightforward and often necessitates a higher number of training iterations.

Apart from that, not many research efforts have been devoted to specialized deep neural architectures for low-dimensional input data, and standard MLPs are generally used by default.
In this paper, we investigate a variant of \ac{rbf} networks \citep{Broomhead1988RadialBF}, which we name Univariate-RBF (\ac{urbf}) layers, to improve trainability for low-dimensional input data. We compare this to existing methods and highlight advantages and limits.
Our approach is inspired by the population coding and the tuning curve stimulus encoding theory from neuroscience~\citep{dayan2005theoretical}.
According to these theories, some neurons encode low-dimensional continuous stimuli by having a bell-shaped tuning curve where their activity peaks at a preferred stimuli value.
\begin{figure}[ht]
    \vskip 0.2in
    \includegraphics[width=\columnwidth]{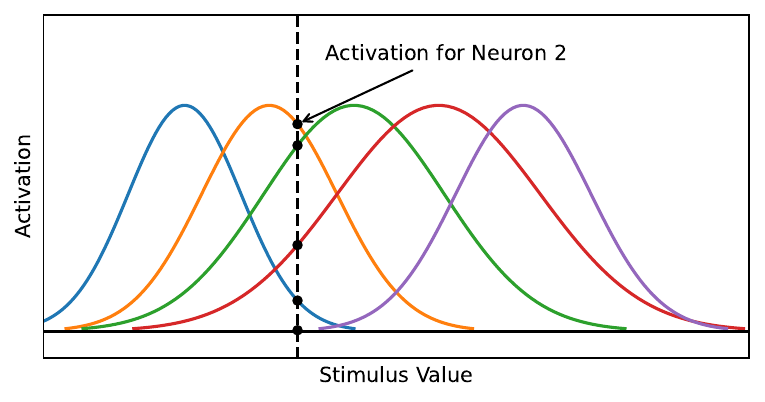} 
    \caption{
    Several neurons encode a single continuous input with Gaussian-shaped activity curves that peak at different receptive values. 
    }
    \label{fig:neural_tuning_curves}
    \vskip -0.2in

\end{figure}
Several neurons (or groups of neurons) encode a single stimulus dimension by having their preferred values span over the stimulus value range (Fig.~\ref{fig:neural_tuning_curves}).
One such example is the processing of movement directions in the medial temporal visual area~\citep{maunsell1983functional} where different populations of neurons are responsive to different directions of movement.
Similar to their neuroscientific inspiration a unit in a \ac{urbf} layer encodes a single continuous input ($x \in \mathbb{R}$) with several neurons ($y \in \mathbb{R}^n$) each having a Gaussian activation function peaking at a given input value $\mu$ with spread $\sigma$. 
The preferred input value ($\mu$) is different for each of the $n$ \ac{urbf} neurons (and possibly their spread too).
These two parameters are learnable for each neuron individually using backpropagation, recalling the adaptability of monkeys' sensory neurons when trained on specific tasks~\citep{schoups2001practising}.



Our contributions in this paper are two-fold:
\begin{itemize}
    \item First, we introduce a new neural network layer, called \ac{urbf}, for function approximation with low-dimensional inputs and prove its universal function approximation capability.
    \item Second, we show the efficiency of the \ac{urbf} layer in low-dimensional function approximation domains and compare it to \ac{ffn} and other state-of-the-art methods
\end{itemize}

\section{Background \& Related Work}

\subsection{Function Approximation}
Function approximation aims to represent potentially complex and unknown functions $y = f(x)$ ($x\in \mathbb{R}^n$, $y \in \mathbb{R}^m$) by a model $y = \tilde{f}_\theta(x)$.
Models are often based on parameters $\theta \in \mathbb{R}^d$.
Many classical methods have been investigated to solve this problem ranging from linear regression \citep{buck1960method}, polynomial regression \citep{Stigler1974Gergonnes1P}, random forests \citep{Breiman2001RandomF}, AdaBoost \citep{Freund1997ADG}, GradientBoosting \citep{friedmanGreedyFunctionApproximation2001}, to support vector machines \citep{Drucker1996SupportVR}.
However, these methods often struggle with high-dimensional input data such as images.
For these cases, deep neural networks (DNNs)~\citep{goodfellow2016deep} such as CNNs~\citep{lecun1989backpropagation,ciresan2011flexible} started to outperform these methods by allowing their training with large amounts of parameters through GPUs.
As a result, DNNs became the research and industry standard for high-dimensional data.
On the other hand, for low-dimensional data, DNNs often reach only a similar performance to traditional methods~\citep{Korotcov2017ComparisonOD,Lee2010ComparisonOS,Jiao2019DoesDL}.
However, the relative ease with which neural networks can be adapted to different tasks and the abundance of tools for their development make them an interesting choice also for low-dimensional problems. 
Increasing their efficacy for such problems benefits the larger machine-learning community.

\subsection{Fourier Feature Mapping}
\label{subsec:ffn}
An important step has been made in \citep{tancik2020fourier} by mapping the input to Fourier features before further processing them using a standard \ac{mlp}. They address a significant challenge in the field of computer vision and graphics: enabling \ac{mlp} to learn high-frequency functions in low-dimensional problem domains. This work is particularly relevant for applications involving the representation of complex 3D objects and scenes using \ac{mlp}s. The authors identify the inherent difficulty of standard \ac{mlp}s in learning high-frequency functions, a phenomenon known as "spectral bias" which has been shown in \citep{biettiInductiveBiasNeural2019}. To overcome this, they propose a novel approach involving a simple Fourier feature mapping according to:

\begin{align*}
\gamma(\mathbf{v}) = [&a_1 \cos{(2\pi\mathbf{b}_1^T\mathbf{v})}, a_1 \sin{(2\pi\mathbf{b}_1^T\mathbf{v})}, \\
&a_2 \cos{(2\pi\mathbf{b}_2^T\mathbf{v})}, a_2 \sin{(2\pi\mathbf{b}_2^T\mathbf{v})}, \\
&\makebox[\widthof{$a_m \cos{(2\pi\mathbf{b}_m^T\mathbf{v})}, a_m \sin{(2\pi\mathbf{b}_m^T\mathbf{v})}$}][c]{\vdots} \\ 
&a_m \cos{(2\pi\mathbf{b}_m^T\mathbf{v})}, a_m \sin{(2\pi\mathbf{b}_m^T\mathbf{v})}]
\end{align*}

which is mapping the input data into Fourier features representing different frequencies, enabling the following \ac{mlp} to learn high-frequency content more effectively. The $\mathbf{b}_m$ vectors are sampled from a Gaussian distribution and represent the direction and frequency for each Fourier feature while $a_m$ is a learnable parameter that is part of the following \ac{mlp}. This approach is significantly enhancing \ac{mlp} performance in relevant low-dimensional regression tasks in computer vision and graphics. This idea is the isotropic generalization of the positional encoding which has already been found to be beneficial in \citep{mildenhallNeRFRepresentingScenes2020, zhongReconstructingContinuousDistributions2020}. To avoid aliasing effects by using Fourier features representing very high frequencies, a cut-off frequency is introduced as an additional problem-specific hyperparameter. This hyperparameter has to be set correctly to achieve competitive results but can be determined beforehand introducing additional complexity to a problem. In the following, we refer to this approach as \ac{ffn} while we also look at a univariate variant which we refer to as \ac{uffn}.


\subsection{RBF Networks}
RBF networks have been used for function approximation and provide the basis for the proposed \ac{urbf} layer.
RBF networks are feedforward networks with a hidden layer.
For an input $\bm{x} \in \mathbb{R}^n$, their $j^{\textrm{th}}$ output is given by:
\begin{equation}
    f_j (\bm{x}) = \sum_{k = 1}^{K} w_{jk} \mathcal{G}(\lVert \bm{x} - \bm{c}_k \rVert,\sigma_k), \hspace{0.8cm}   
    \label{eq:rbf_network}
\end{equation}
for
\[
j = 1, 2, \ldots , J,
\]

where $\mathcal{G}(u,v)=\exp(-u^2/2v^2)$ is a Gaussian kernel, $J$ represents the number of outputs, $K$ represents the number of Gaussian kernels in the layer, and the $w_{jk}$'s are scalar weights. 
The center and the standard deviation of the $k^{\textrm{th}}$ RBF kernel are represented as $\bm{c}_k\in\mathbb{R}^D$ and $\sigma_k>0$. 

RBF networks were introduced as adaptive networks for multivariable functional interpolation \citep{Broomhead1988RadialBF} and have seen widespread adaptation since Park and Sandberg  \citep{Park1991UniversalAU, Park1993ApproximationAR} proved that they are universal function approximators, i.e.\ an RBF network with one hidden layer and certain assumptions regarding their kernel functions is capable of approximating any function to a required amount of accuracy.
The kernel function plays an important role, and several kernel functions have been proposed \citep{Montazer2018RadialBF}.
However, the Gaussian kernel is the most widely used. 
An important characteristic of a Gaussian kernel is that its activation decreases monotonically with the distance from its center point $\bm{c}_k$, with the decrease in the magnitude controlled by its spread $\sigma_k$, often termed as width.
This property allows them to act as local approximators.

Considerable research has been dedicated to the selection of centers for Gaussian kernels \citep{Gan1996DesignFC,Perfetti2006ReducedCR,Warwick1995CentreSF,Burdsall1997GARBFAS,Bhatt2004OnTS}. 
A naive approach is to randomly choose the centers of RBF, suggesting that it helps with regularization \citep{kubat1999neural}. 
However, such approaches are unsustainable for large datasets. 
Meyer et al. \citep{Meyer2017NearestNR} developed a widely used method to choose RBF centers by using clustering algorithms to determine the cluster patterns in the input and select a sample from each cluster as a center for RBF neurons. 
Our work treats the centers as parameters and learns them using gradient updates.

Different variants of RBF networks have seen wide use in supervised learning tasks, involving application areas such as biology/medicine \citep{Venkatesan2006ApplicationOA,Liu2022AMR,Liu2021IntelligentAP} and control systems \citep{Gu2008FuzzyRB,Tan2005AdaptiveOC}.
One of the more closely related works to our work is from Jiang et al. \citep{Jiang2021AnEM}.
They propose a multilayer RBF-MLP network in which they replace the activation function with Univariate RBF in between MLP layers this deviates from our approach where we use the \ac{urbf} layer only as the first layer and map the lower dimensional input data into a higher dimensional space. 

\section{Univariate Radial Basis Function Layer}
\label{c:urbf}
\begin{figure*}[t]
\vskip 0.2in
  \centering
    \includegraphics[width=1.4\columnwidth]{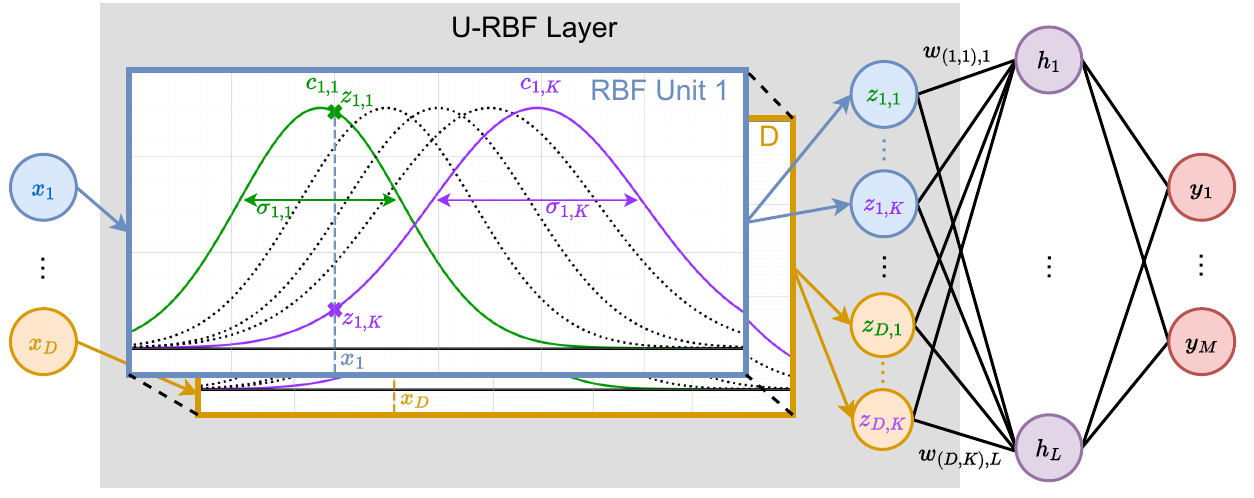}
    \caption{The \ac{urbf} layer.}
  \label{fig:urbf_network}
  \vskip -0.2in
\end{figure*}

This section introduces the univariate radial basis function layer (Fig.~\ref{fig:urbf_network}) which is derived from RBF networks (\ref{eq:rbf_network}).
From here on, the classical RBF network structure will be referred to as \ac{mrbf} layer as it uses multivariate Gaussian kernels.
In contrast, the univariate layer uses 1-dimensional Gaussian kernels.

The \ac{urbf} layer consists of $D$ RBF units, one for each dimension of the input $\bm{x}\in\mathbb{R}^D$.
The individual input dimensions $x_d$ are remapped into a higher-dimensional representation using $K_d$ 1-dimensional Gaussian kernels per RBF unit. 
The layer's output dimension is $J=\sum_{d=1}^D K_d$, where $J=KD$ if the number of kernels is the same for all RBF units.
This high-dimensional representation is fed into a fully connected layer with $L$ output neurons, indexed by $l$:
\begin{equation} h_l(\bm{x}) = \sum_{d = 1}^{D} \sum_{k = 1}^{K_d} w_{(d, k), l}~z_{d,k} \end{equation} 
with
\[z_{d,k} = \mathcal{G}(x_d - c_{d, k}, \sigma_{d, k})~,\]
where the high-dimensional representation is denoted by $z_{d,k}$ with a double index, to map it to each of the $D$ input dimensions (RBF units) and $K_d$ Gaussian kernels.
The centers $c_{d,k}$ (and potentially the spreads $\sigma_{d,k}$) per RBF unit are different so that their activation differs over the values inside the value range of their input dimension $x_d$.
The centers and spreads can be either defined or learned by gradient descent with backpropagation together with the weights $w_{(d,k),l}$.
The number of Gaussian kernels per input (RBF unit) is a hyperparameter named as Number of Neurons Per Input (NNPI). 
In summary, the \ac{urbf} layer projects each dimension of an input to a higher dimensional space via its RBF units, where the projection depends on the centers and spreads of its Gaussian kernels. 


\subsection{Universal Function Approximation}
\label{c:app_theory}
A \ac{urbf} layer that is followed by a hidden layer is a Universal Function Approximator, i.e.\ it can accurately approximate any complex function to an arbitrary degree of precision. 
More formally:
\begin{theorem}
Let $\{\bm{x}_1, \bm{x}_2, \dots, \bm{x}_N\}$ be a set of distinct points in $\mathbb{R}^D$ and let $f : \mathbb{R}^D \to  \mathbb{R}^L$ be any arbitrary function. 
Then there is a function $g: \mathbb{R}^D \to \mathbb{R}^L$ with \ac{urbf} layer with $K \ge 2$ Gaussian kernels with different kernels, such that $f(\bm{x}) = g(\bm{x})$, for all $x \in \{\bm{x}_1, \bm{x}_2, \dots, \bm{x}_N\}$.
\label{urbf_univ}
\end{theorem}
The proof is given in Appendix~\ref{sec:appendix_proofs}. 
The theorem provides the theoretical justification that \ac{urbf} networks, similar to MLPs, have the representation power to capture any functional relationship, given sufficient parameters. 
This is a fundamental requirement for the network to be applicable across different tasks and problem domains.

\section{Experiments}
\label{exp}

In this section, we describe the experiments conducted to evaluate the performance and behavior of the novel \ac{urbf} layer in comparison to existing deep learning approaches, including \ac{mrbf}, \ac{ffn}, and \ac{mlp} and traditional approaches such as \ac{svr} and Gradient boosting. The objective of these experiments is to analyze \ac{urbf}'s and the other method's properties in terms of their ability to regress different problems. More information about the experiments can be found in Appendix ~\ref{app:experiment_details}.

\subsection{White Noise Regression}
\begin{figure}[b!]
  \centering
    \includegraphics[width=0.7\columnwidth]{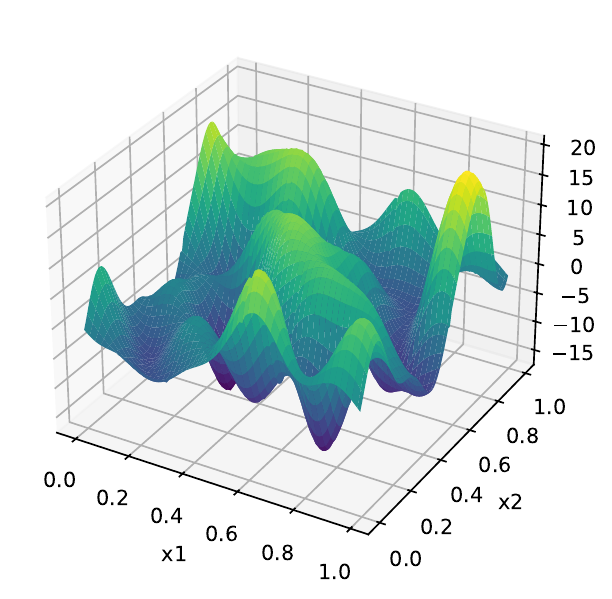}
    \caption{Sample of isotropic low-pass filtered White Noise which serves as the target function. The cut-off frequency is set to $4 Hz$.\label{fig:white_noise}}
\end{figure}

We explore the capabilities of the investigated methods by regressing isotropic low-pass filtered white noise. It does not contain an axis-aligned bias and is therefore appropriate to evaluate our approach. A sample target function is visualized in Fig.~\ref{fig:white_noise} and the details about the generation of isotropic low-pass filtered white noise can be found in Appendix ~\ref{app:white_noise_gen}. We first assess the performance of a large variety of methods on a single white noise regression problem. Here we look at different initialization methods for the parameters of the \ac{rbf} Neurons. Furthermore, for \ac{ffn} we look at \ac{uffn} variant where we compare two different initialization methods. The uniform initialization uses evenly log-linear spaced frequencies which is equal to the positional encoding described in \ref{subsec:ffn}. In addition, we evaluate the usage of a random initialization where we sample the frequencies based on a Gaussian distribution as in the original \ac{ffn} implementation. However, there's a key difference: we set only one randomly selected input dimension to a non-zero value. This adjustment turns the method into a univariate approach. 


\begin{figure}[b!]
  \centering
    \includegraphics[width=\columnwidth]{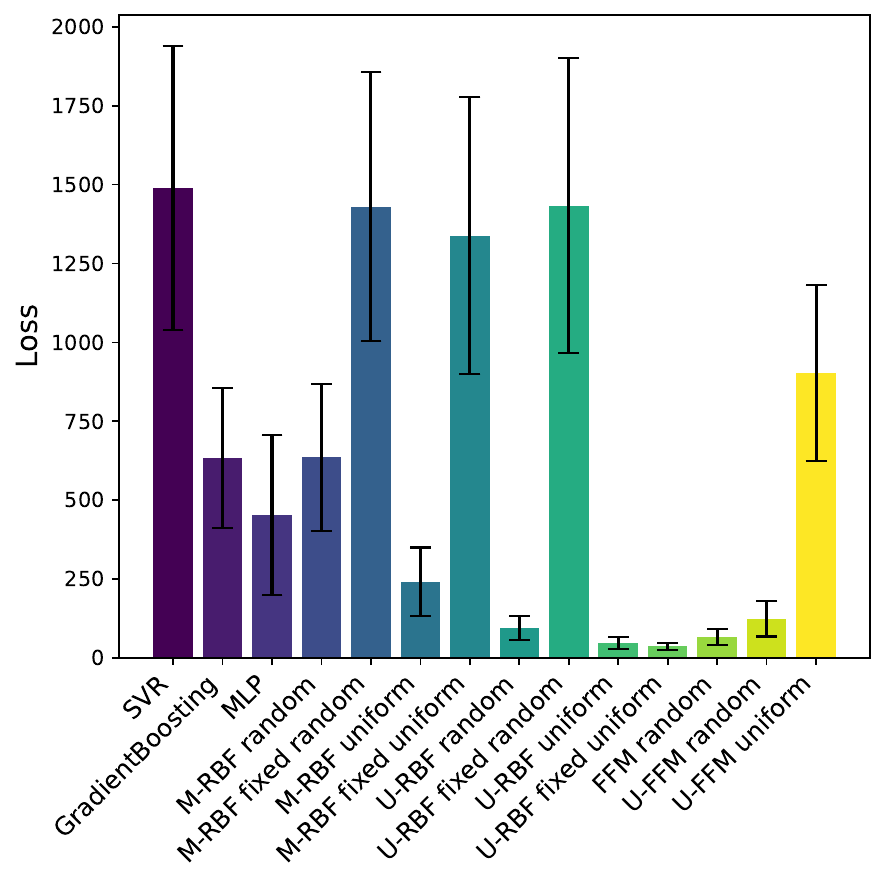}
    \caption{The U-RBF outperforms competing approaches across different models for white noise regression. The term 'fixed' refers to non-learnable RBF Neurons while 'random' and 'uniform' refer to the initialization strategies.\label{fig:white_noise_base}}
\end{figure}

 The results are depicted in Fig.~\ref{fig:white_noise_base} and show a huge difference in loss for the different approaches. For the \ac{mrbf} model family, the combination of a uniform initialization of the \ac{rbf} input neurons while using learnable \ac{rbf} parameters performs best. For the \ac{urbf} model family a similar observation can be made with the difference that the non-learnable \ac{urbf} layer performs slightly better than the learnable. Therefore, in the following experiments, we further focus on the learnable and non-learnable \ac{mrbf} and \ac{urbf} variation with uniform initialization of the \ac{rbf} neurons. The \ac{ffn} approach results are comparable to the \ac{urbf} results and will be further investigated. The \ac{uffn} model shows good results when using random initialization and will be therefore further examined in the following.

\begin{figure}[t]
  \centering
    \includegraphics[width=1\columnwidth]{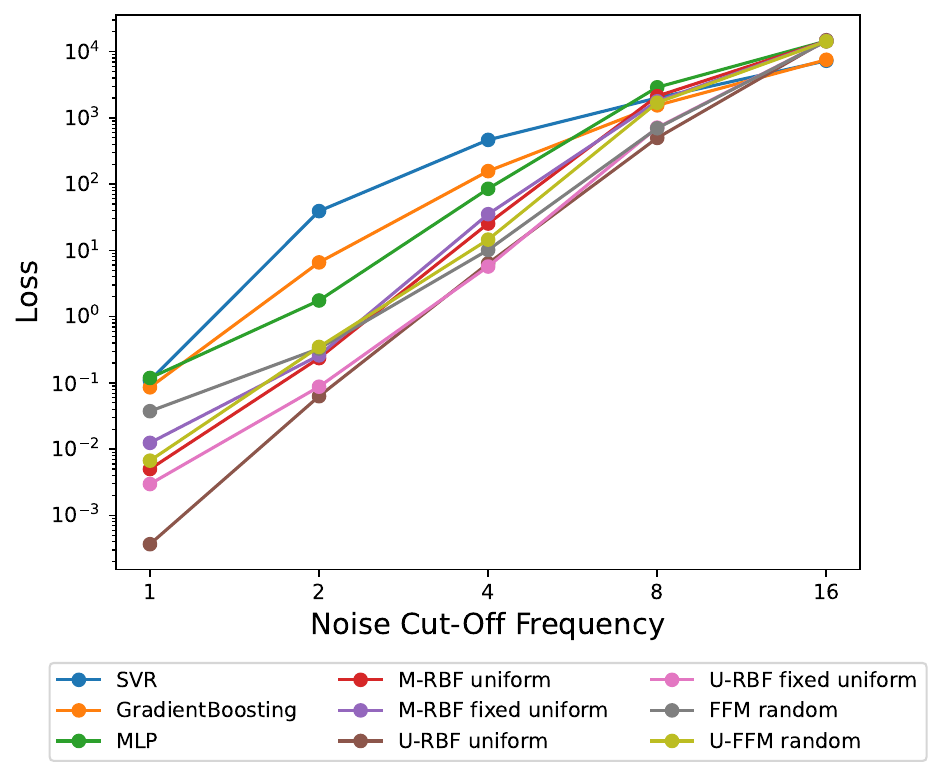}
    \caption{Performance across different noise cut-off frequencies. The \ac{urbf} approach shows the best performance for the lower four frequencies. \label{fig:frequency_sweep}}
\end{figure}
\begin{figure}[t]
  \centering
    \includegraphics[width=1\columnwidth]{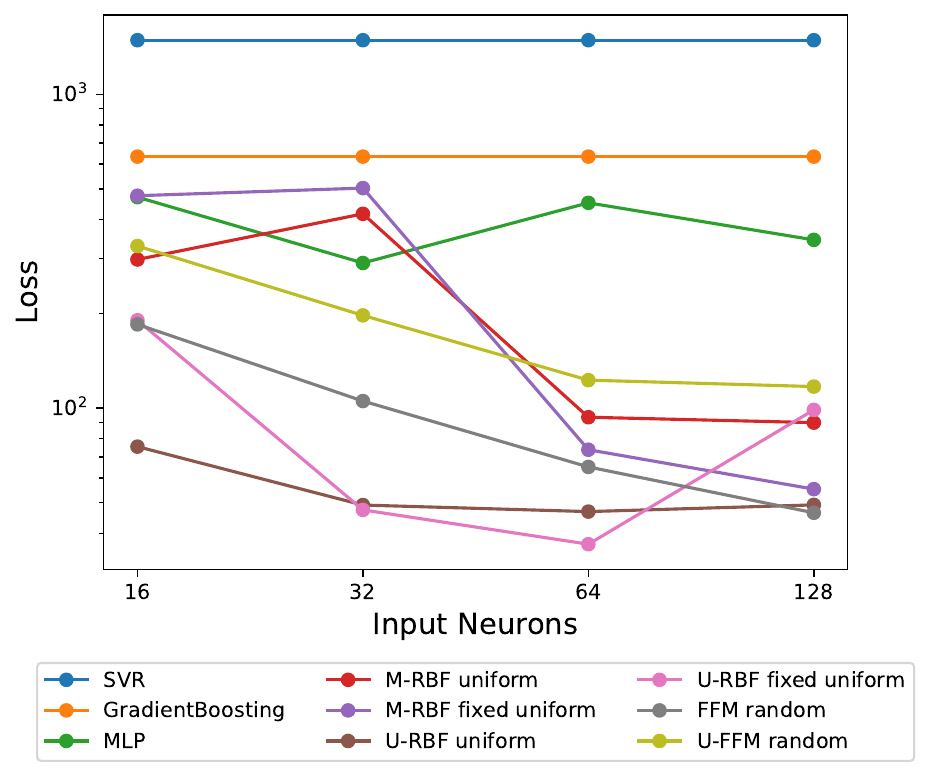}
    \caption{Performance across different numbers of input neurons. While \ac{ffn} is advantageous for a high number of input neurons, \ac{urbf} shows strong results across all numbers of input neurons.\label{fig:neuron_sweep}}
\end{figure}
\begin{figure}[t]
  \centering
    \includegraphics[width=1\columnwidth]{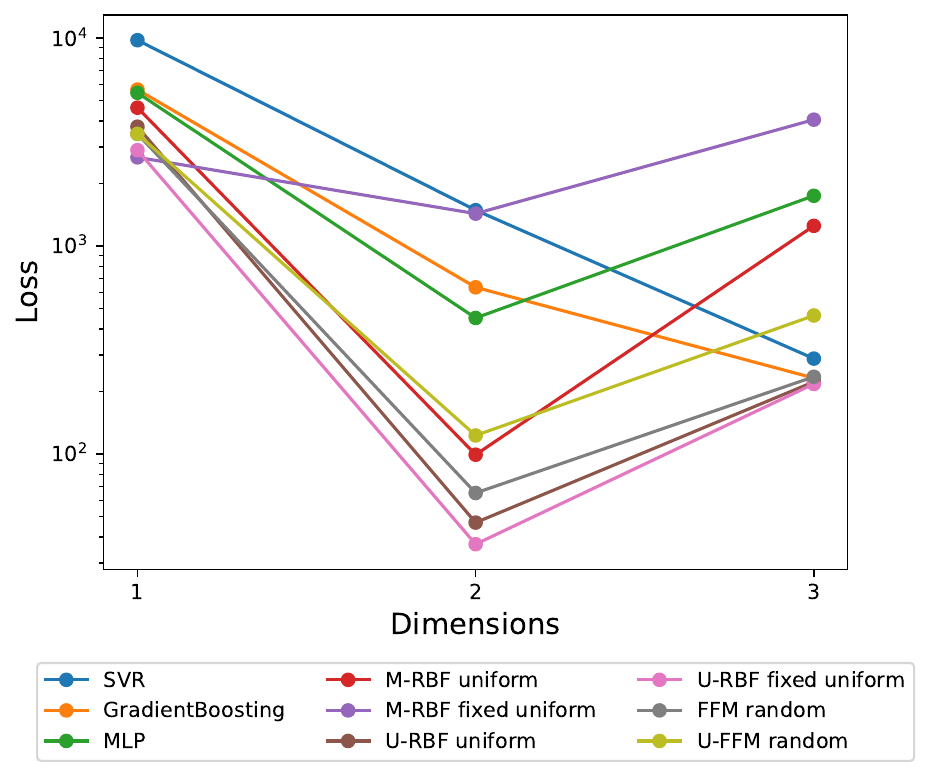}
    \caption{Performance across different noise input dimensions. \ac{rbf} based approaches perform best across all settings. \ac{urbf} shows superior results for the two higher input dimensions. \label{fig:dimension_sweep}}
\end{figure}

\begin{figure}[t]
  \centering
    \includegraphics[width=1\columnwidth]{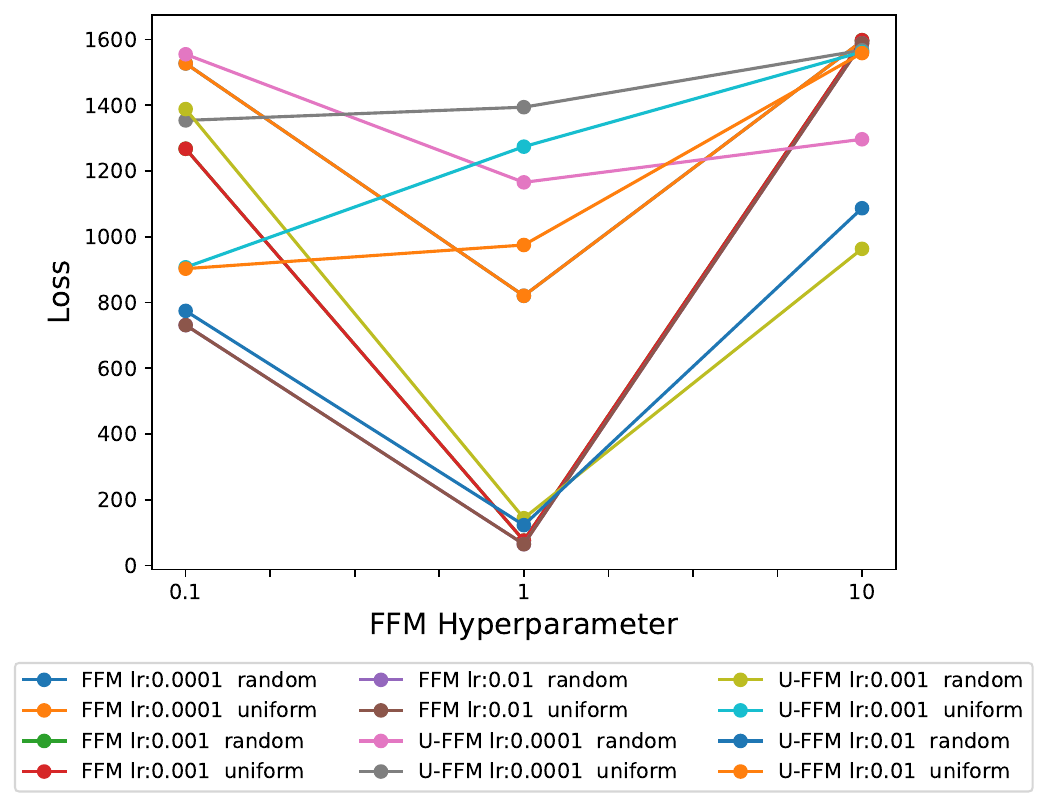}
    \caption{Performance across different \ac{ffn} hyperparameters. Even though the target function is the same for all variants, the optimal \ac{ffn} hyperparameter differs suggesting a complex dependency. \label{fig:ffm_hyperparameter_sweep}}
\end{figure}

\subsubsection{Frequency Sweep}
We investigate the low-pass filtered white noise regression performance across different cut-off frequencies in Fig.~\ref{fig:frequency_sweep}. Across all frequencies, a clear behavior can be observed since the performance of all methods decreases with increasing cut-off frequency. The \ac{urbf} model shows the best performance across all cut-off frequencies except the highest where the traditional methods, specifically \ac{svr} and GradientBoosting perform best.

\subsubsection{Input Neuron Sweep}
The number of input neurons is significant for handling complex regression tasks, especially for the \ac{ffn} approach and the \ac{rbf} approaches. Here we set the number of input dimensions to $2$ and the cut-off frequency to $4$ Hz. The results are depicted in Fig.~\ref{fig:neuron_sweep} where the increasing number of neurons show a decrease in the loss for the \ac{ffn}, \ac{urbf}, and \ac{mrbf} approaches. The \ac{urbf} approach has the lowest loss across all numbers of input neurons except for $128$ neurons. The \ac{ffn}-based approach has a slightly smaller loss in this case.

\subsubsection{Dimension Sweep}
For different numbers of dimensions, the behavior is depicted in Fig.~\ref{fig:dimension_sweep} where the \ac{urbf} and the \ac{ffn} approach show the lowest loss. Especially in the two-dimensional case, the is clear compared to the other methods. For one-dimensional input, the performance advantage moves from the \ac{urbf} approach towards the \ac{mrbf} approach which performs slightly better in this case.

\subsubsection{FFM Hyperparameter Sweep}
The \ac{ffn}-based approaches require defining a cut-off frequency to generate Fourier feature mappings in the right frequency range. The results for different hyperparameter settings are visualized in Fig.~\ref{fig:ffm_hyperparameter_sweep} and show that the correct setting of this hyperparameter is crucial to achieving competitive results. Nonoptimal hyperparameter settings have a much higher loss. Furthermore, it is observable that some variants require different hyperparameter settings since the loss is not minimal for the parameter of $1$ in these cases, suggesting a complex dependency between the optimal hyperparameter, the problem setting, and the \ac{ffn}-based method. The influence of the hyperparameter settings for different problem settings can be found in Appendix ~\ref{app:hyperparameter_settings}.

\begin{figure}[b]
  \centering
    \includegraphics[width=0.8\columnwidth]{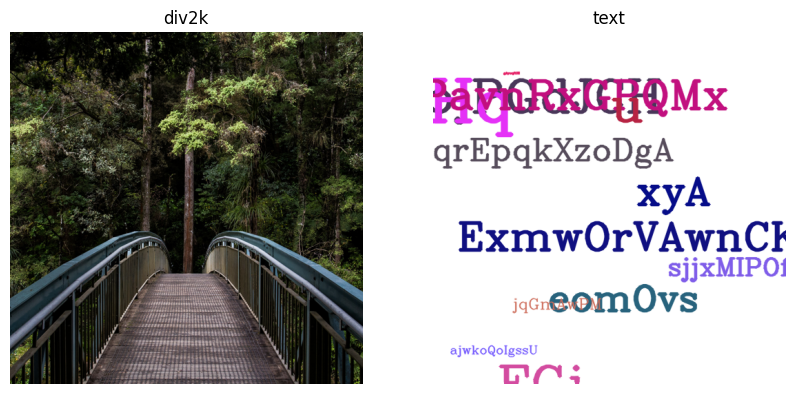}
    \caption{Figures for the image regression task. While one contains a complex and detailed scene, the other is mostly homogenous and contains sparse structures imposing two very different problem settings. \label{fig:image_regression_targets}}
\end{figure}

\subsection{Image Regression}
In \cite{tancik2020fourier}, an image regression task is conducted to demonstrate the effectiveness of their proposed \ac{ffn} in enhancing the ability of \ac{mlp} to learn high-frequency functions. In the following, we will replicate this experiment to compare the performance of the introduced \ac{urbf} Layer. As in \cite{tancik2020fourier}, the task is to learn a function that maps the input coordinates to the corresponding RGB values of the target image. The training set consists of linearly spaced points, with every \(c\)'th point used for training and the remaining points for testing. We compare all methods on two separate images. Both images are visualized in Fig.~\ref{fig:image_regression_targets} and show fundamentally different contents. While one contains a complex and detailed scene, the other is mostly homogenous and contains sparse structures imposing two very different problem settings.

\begin{figure}[t]
  \centering
    \includegraphics[width=\columnwidth]{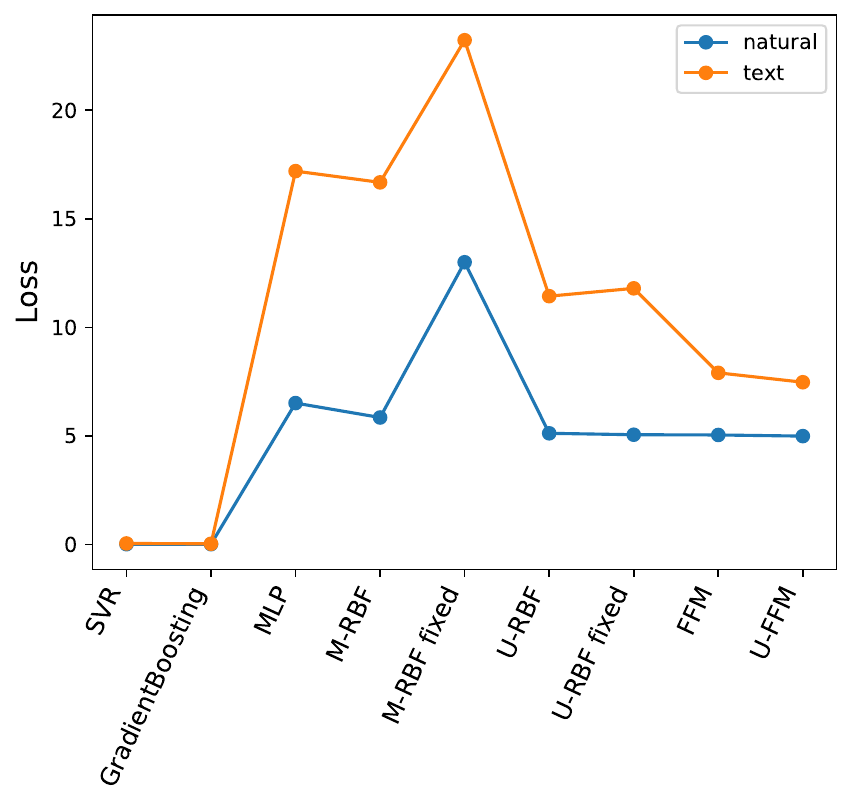}
    \caption{Loss for each target image per used method. The traditional methods perform best on this task but can not be directly compared to the deep learning models in terms of model size and complexity. Among the deep learning approaches the \ac{ffn} based approaches perform best especially for the image containing sparse structures. \label{fig:image_regression}}
\end{figure}

For the image regression task the \ac{ffn} approach has a smaller loss for both target images compared to other used deep learning models. This difference is especially strong for the second image where the \ac{ffn} approach is surpassed by the \ac{uffn} model. Apart from that, the traditional methods have a clear advantage here and show superior performance as visualized in Fig.~\ref{fig:image_regression}.

\subsection{Real World Dataset Regression}
\begin{figure}[t]
  \centering
    \includegraphics[width=\columnwidth]{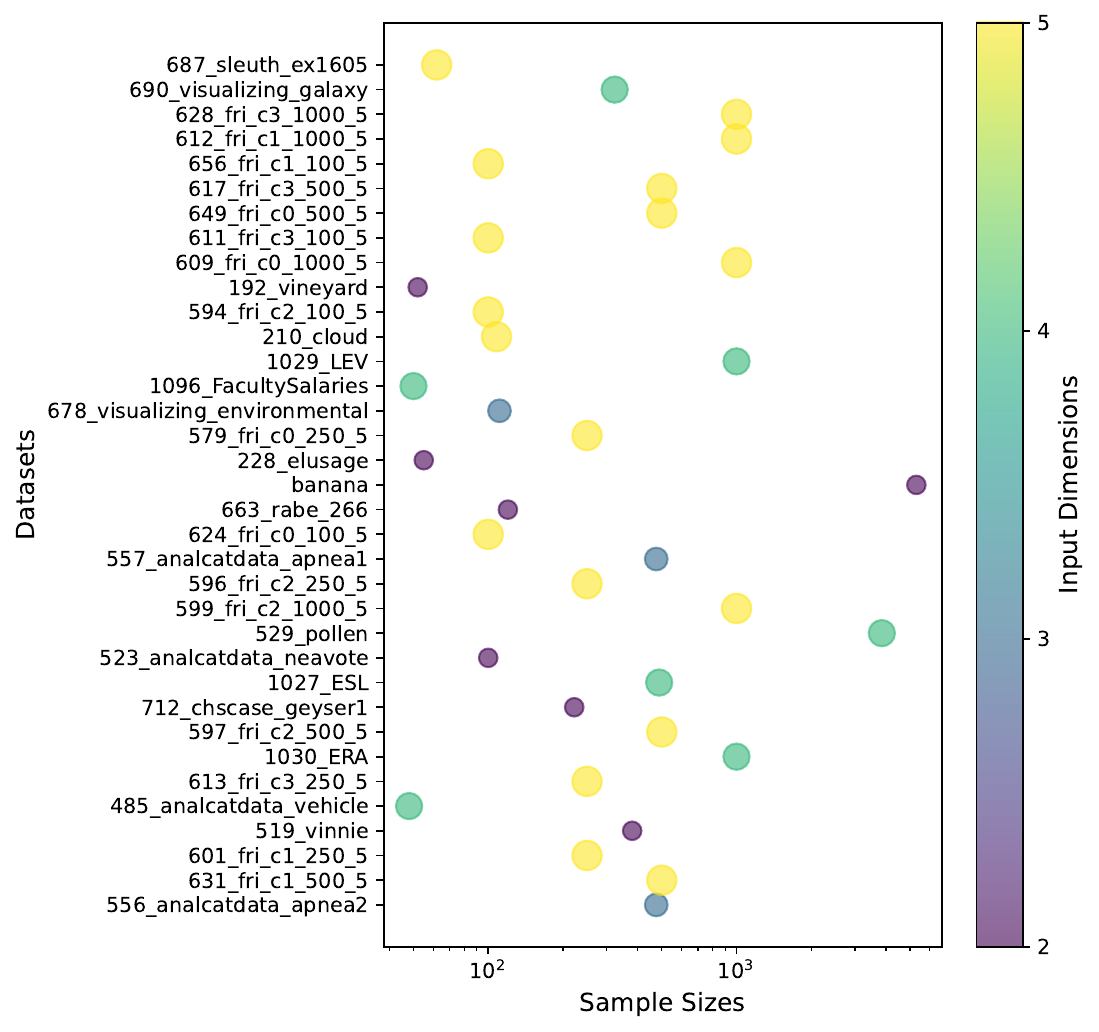}
    \caption{Input dimensionality and sample size of the used real-world datasets. Most of the 5-dimensional datasets seem to originate from the same base data suggesting a similar data structure.\label{fig:datasets_summary}}
\end{figure}

To prove the effectiveness on real-world problems, we evaluate the models on low dimensional regression tasks from \ac{pmlb}. We only select datasets with an input dimensionality of not larger than $5$. The datasets are visualized in Fig.~\ref{fig:datasets_summary} by input dimensionality and sample size. To be able to compare the results between datasets better, we divide the results for each dataset by the lowest loss. These adjusted results can then be used to average across datasets to grasp the broader picture. 

\begin{figure}[t]
  \centering
    \includegraphics[width=\columnwidth]{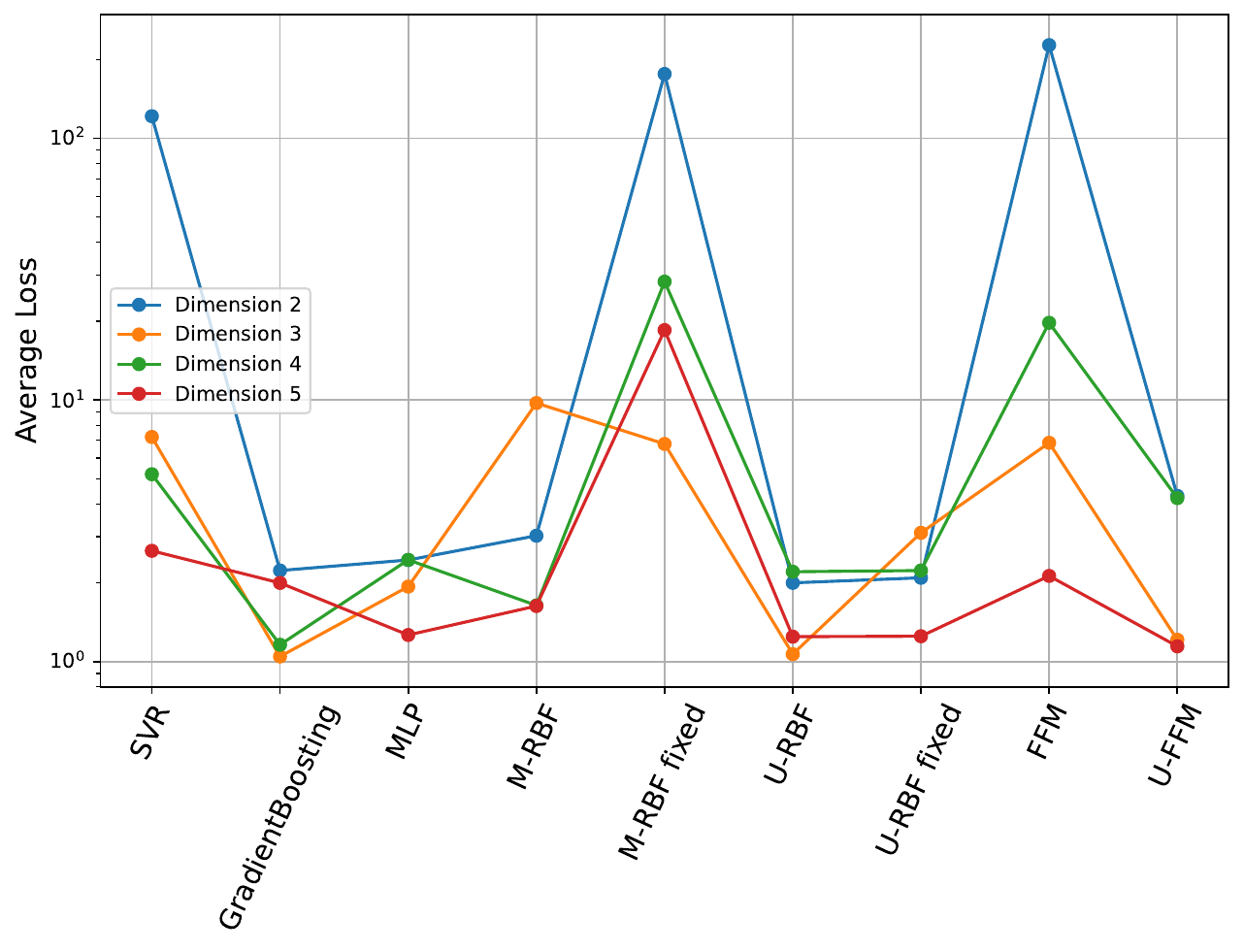}
    \caption{Average Loss per Dataset Input dimension. Among the deep learning based methods, the \ac{urbf} approach achieves overall competitive results across all input dimensionalities. \label{fig:full_dataset_avr_per_dim}}
\end{figure}

The results are grouped depending on the input dimension of the dataset and depicted in Fig.~\ref{fig:full_dataset_avr_per_dim}. There is no clear superiority observable. Among the deep learning methods, the \ac{urbf} approach has the best overall performance which is observable in \ref{tab:full_dataset_avr_per_dim}. The complete evaluation results can be found in Appendix \ref{app:dataset_regression}.

s

\begin{table*}[t]
\centering
\caption{Average scaled loss per dataset input dimension. Methods with the term 'Fixed' refer to variants with non-learnable \ac{rbf} units.\label{tab:full_dataset_avr_per_dim}}
\begin{tabular}{c|cccccccccc} 
\toprule
 \#dim & SVR & GradBoost & MLP & M-RBF & M-RBF Fixed & U-RBF & U-RBF Fixed & FFN & U-FFN \\
\midrule
2 & 121.36 & 2.23 & 2.44 & 3.03 & 175.99 & \textbf{2.00} & 2.09 & 226.85 & 4.30 \\
3 & 7.21 & \textbf{1.05} & 1.93 & 9.73 & 6.79 & 1.07 & 3.10 & 6.85 & 1.21 \\
4 & 5.20 & \textbf{1.16} & 2.45 & 1.64 & 28.31 & 2.20 & 2.23 & 19.71 & 4.21 \\
5 & 2.65 & 2.00 & 1.26 & 1.63 & 18.48 & 1.25 & 1.25 & 2.12 & \textbf{1.14} \\
\bottomrule
\end{tabular}
\end{table*}

\section{Discussion}
Even though the \ac{urbf} layer imposes a bias towards changes along axial directions, we can show its superiority for problems that do not benefit from this bias. Especially for settings with a limited amount of input neurons and lower target function frequencies, the \ac{urbf} layer is able to outperform other approaches by a large margin.

In the context of image regression, the superiority of the \ac{urbf} layer is not given. The traditional approaches are superior in this problem setting because of simple reasons. In the case of \ac{svr}, the method is designed to store a collection of support vectors up to the size of the training dataset making this approach only applicable to problems with a fairly small amount of training samples. Since the training samples are equidistant sampled points, this problem can be solved fairly easily using the stored support vectors and is therefore not entirely comparable to the deep learning approach. The same holds for the Gradient Boosting approach where the focus is on sequentially correcting the errors of the previous models, making it highly adaptable and efficient for structured datasets. However, like SVR, its effectiveness may diminish with extremely large and complex datasets where deep learning methods might have an advantage due to their ability to automatically extract and learn high-level features from raw data.

In the domain of real-world dataset regression, the varying performance of the models highlights the importance of considering the specific characteristics of the data when choosing an appropriate predictive model. Nevertheless, the \ac{urbf} shows the best overall performance among the deep learning methods therefore suggesting robust properties enabling to adapt to different datasets. 

Across all experiments, the \ac{urbf} is capable of showing comparable or superior performance to existing deep learning models and traditional approaches. In comparison to \ac{ffn} based approaches, alternative explored strategies demonstrate the advantage of not requiring the identification of an optimal cut-off frequency hyperparameter to yield satisfactory outcomes. This hyperparameter has a huge impact on the results and can not be easily determined in advance. It is necessary to try different values in order to find an appropriate hyperparameter setting, introducing a considerable overhead and rendering this approach less practical in most scenarios. In contrast to that, for \ac{rbf}-based approaches, the range of the input data has to be known for the initialization process which is usually the case for supervised learning tasks, where the whole dataset is accessible in advance. 

Furthermore, we study the impact of the axis directional bias in the methods by investigating also the \ac{uffn} approach which is closely related to the positional encoding. In the case of real-world datasets, this axial bias seems advantageous since it aligns with the dataset but apart from that the univariate property in the \ac{urbf} layer might also instruct a more simple learnable framework compared to the investigated \ac{mrbf} approach while introducing the beneficial local properties of \ac{mrbf}. 

\section{Conclusion}

In conclusion, the introduction of the \ac{urbf} layer in the context of regression tasks has demonstrated a significant improvement in performance over commonly used methods. The \ac{urbf} layer's inherent bias towards axial changes does not hinder its effectiveness, even in scenarios where this characteristic is not specifically advantageous. Particularly in cases with fewer input neurons and lower target function frequencies, the \ac{urbf} layer outperforms competing approaches considerably. In the context of datasets with low dimensional inputs the \ac{urbf} layer improves the results compared to other methods such as \ac{ffn} or standard \ac{mlp}. The results suggest that the \ac{urbf} introduces beneficial properties in terms of a simple-to-learn framework that can easily adapt to the given data compared to \ac{mrbf}, especially with a limited amount of neurons. In contrast to the investigated \ac{ffn}-based approach the \ac{urbf} method does not include sweeping through complex hyperparameter settings. Only the range of the input data has to be known in advance.
Overall, the \ac{urbf} layer's introduction marks a considerable advancement in the field of regression analysis, offering robustness, adaptability, and often superior performance compared to both traditional and other deep learning methods. This demonstrates its potential as a valuable tool in various regression tasks, particularly where data characteristics are well-aligned with its strengths.

\section*{Broader Impact}
This paper presents work whose goal is to advance the field of Machine Learning. There are many potential societal consequences of our work, none which we feel must be specifically highlighted here.

\section*{Funding}
This research was partially supported by the SPRING project funded by the European Commission under the Horizon 2020 framework programme for Research and Innovation (H2020-ICT-2019-2, GA \#871245) and by the the project ML3RI funded by the Young Researcher (Jeunes Chercheuses et Jeunes Chercheurs) programme of the ANR, under grant agreement ANR-19-CE33-0008-01.
\nocite{langley00}

\bibliography{references}
\bibliographystyle{icml2024}

\newpage
\appendix
\onecolumn

\section{Theoretical Results}
\label{sec:appendix_proofs}
A critical theoretical foundation for the proposed U-RBF network architecture is establishing its universal function approximation capability. This section provides the formal analysis to prove U-RBF networks can approximate any continuous function over the input domain.
We first define the function classes used to represent single hidden layer feedforward neural networks (Definitions \ref{def:affine} and \ref{def:ff_network}). Theorem \ref{theo:mlp_universal_approx}, adapted from \citep{Hornik1989MultilayerFN}, establishes the universal approximation property for this network class. We then prove a one-to-one mapping between the input space and U-RBF layer output space in Corollary \ref{cor:two}. Finally, Theorem \ref{urbf_univ} leverages these results to formally demonstrate that augmenting the single hidden layer networks with a U-RBF layer maintains the universal function approximation property.

\begin{definition}
    For $J \in \mathbb{N}$, $\mathcal{A}^J$ is defined as the set of affine functions from $\mathbb{R}^J$ to $\mathbb{R}$, where an affine function is of the form $A(\bm{z}) = \bm{w}^\top \bm{z} + b$, $\bm{w},\bm{z}\in\mathbb{R}^J$ and $b\in\mathbb{R}$.
    \label{def:affine}
    \end{definition}
In the context of feedforward networks or MLPs, the affine function is the linear transformation of the input $x$ using a weight vector plus an offset or bias. 
A non-linear transformation following an affine function is widely used in  machine learning models. 
Thus we introduce one such class of functions in the next definition based on our current definition of affine functions. 
\begin{definition}
    For any measurable function $\rho(\cdot)$ from $\mathbb{R}$ to $\mathbb{R}$ and $J \in \mathbb{N}$ let $\mathcal{B}^J(\rho)$ be the set of functions defined as:
    $$
    \mathcal{B}^J(\rho) = \left\{f: \mathbb{R}^J \rightarrow \mathbb{R},~ f(\bm{z}) 
    =\sum_{q=1}^Q \beta_q \rho\left(A_q(\bm{z})\right), \beta_q \in \mathbb{R}, A_q \in \mathcal{A}^{J}, 
    Q\in\mathbb{N} \right\} .
    $$ \label{def:ff_network}
\end{definition}
This class of functions is used to represent single-layer feed-forward networks with input space $\mathbb{R}^J$ and output space $\R$. 
The $\beta_q$'s represent scalar weights from the hidden layer to the output layer. Further from now, we will use this class of functions to represent our single-layered single-output feed-forward network. 


\begin{theorem}
Let $\left\{\bm{z}_{1}, \ldots, \bm{z}_{N}\right\}$ be a set of distinct points in $\mathbb{R}^{J}$ and let $g: \mathbb{R}^{J} \rightarrow \mathbb{R}$ be an arbitrary function. If $\psi$ achieves 0 and 1 , then there is a function $f \in \mathcal{B}^{J}(\psi)$ with $N$ hidden units such that $f\left(\bm{z}_{i}\right)=g\left(\bm{z}_{n}\right), n \in\{1, \ldots, N\}$.
\label{theo:mlp_universal_approx}
\end{theorem}
Theorem~\ref{theo:mlp_universal_approx} taken from \citep{Hornik1989MultilayerFN} proves the universal function approximation capability of feedforward neural networks. Although we only state the theorem for a single hidden layered network with a single output, the authors in the same paper generalize it to a multi-layer multi-output network. With the universal function approximation capability of a general feedforward network proven, we now move on to corollary \ref{cor:two} to prove one-to-one mapping from the input space to the intermediate space, which is the output of our \ac{urbf} layer.
\begin{corollary}\label{cor:two} For any $K\geq 2$, $c_1,\ldots,c_K\in\mathbb{R}$ and $\sigma_1,\ldots,\sigma_K >0$, the mapping $h: \mathbb{R} \rightarrow \mathbb{R}^K$ defined as:
$$
h(u) = \left(\exp\left(-\frac{(u - c_1)^2}{2\sigma_1^2}\right), \exp\left(-\frac{(u - c_2)^2}{2\sigma_2^2}\right), \dots, \exp\left(-\frac{(u - c_K)^2}{2\sigma_K^2}\right) \right)
$$
is a one-to-one correspondence for all $u \in \mathbb{R}$, if and only if $c_i \ne c_j$ for at least one pair of $(i, j)$, $i \ne j$.

\end{corollary}
\textit{Proof. }It is needed to prove that if $h(u) = h(v)$ then $u = v$.
For the sake of simplicity and without loss of generality, let us assume $K = 2$. This implies for $h(u) = h(v)$:
$$
\left(\exp\left(-\frac{(u - c_1)^2}{2\sigma_1^2}\right), \exp\left(-\frac{(u - c_2)^2}{2\sigma_2^2}\right)\right) = \left(\exp\left(-\frac{(v - c_1)^2}{2\sigma_1^2}\right), \exp\left(-\frac{(v - c_2)^2}{2\sigma_2^2}\right)\right)
$$
Since $\exp(\cdot)$ is a monotonically increasing function we have:
\begin{equation}
(u - c_1)^2 = (v - c_1)^2
\quad\text{and}\quad
(u - c_2)^2 = (v - c_2)^2.
\end{equation}
Therefore, given that $c_1 \ne c_2$, the only possible solution 
is $u = v$. Thus, $h$ is a one-to-one correspondence for all $K \ge 2$. 

Corollary~\ref{cor:two} proves one-to-one mapping from the input space to the higher dimensional space that \ac{urbf} projects into. The corollary is essential for the universal approximation property proved in the next theorem, in the sense that it reflects a unique mapping from the input space to the output space for the U-RBF layer and thus provides the basis for further transformation in the following layers.

\textbf{Theorem~\ref{urbf_univ}.}~\textit{Let $\{\bm{x}_1, \bm{x}_2, \dots, \bm{x}_N\}$ be a set of distinct points in $\mathbb{R}^D$ and let $f : \mathbb{R}^D \to  \mathbb{R}^L$ be any arbitrary function. 
Then there is a function $g: \mathbb{R}^D \to \mathbb{R}^L$ with \ac{urbf} layer with $K \ge 2$ Gaussian kernels with different kernels, such that $f(\bm{x}) = g(\bm{x})$, for all $x \in \{\bm{x}_1, \bm{x}_2, \dots, \bm{x}_N\}$.}

\textit{Proof. }From Theorem~\ref{theo:mlp_universal_approx} and Corollary~\ref{cor:two}, it is clear that if there is a one-to-one mapping from input space to a higher dimensional space using \ac{urbf}, following which an MLP is applied to this space, then the network is able to approximate any function $f$. Although the theorem proves the universal approximation for a single output, a set of further tedious modifications in turn would make it applicable for multiple output networks with more than one hidden unit.

As it can be noticed, our approach mainly hinges on the point that there is no loss of information while projecting the input space into a higher dimensional space using the \ac{urbf} layer. 
Using this along with the universal function approximation capability of a feedforward network, we conclude that in turn adding our proposed \ac{urbf} layer to a feedforward network doesn't affect its universal function approximation capability.

\section{Experiment Details}
\label{app:experiment_details}
For every method and hyperparameter setting $4$ repetitions are executed and averaged. Furthermore, we evaluate all learnable models on three learning rates ($0.01$, $0.001$, $0.0001$) and pick the best for each approach. For the \ac{ffn} model, the cut-off frequency hyperparameter is crucial to achieve competitive results. As a result, we evaluate each learning rate on three different cut-off frequencies and pick the best performance. The datasets are split into a train, val, and test part of $0.6$, $0.2$, and $0.2$. The traditional approaches such as Gradient boosting and \ac{svr} are parametrized differently and are therefore not entirely comparable to the deep learning approaches in terms of model size and complexity. The \ac{svr} model uses a \ac{rbf} kernel and stores all support vectors which determine the decision boundary. The amount of stored model parameters is therefore dependent on the size and complexity of training data. The gradient boosting approach performs $100$ boosting stages to fit the data.

\section{White Noise Generation}
\label{app:white_noise_gen}
The core of the evaluation is the regression of n-dimensional Gaussian white noise, denoted as \( f_{Noise}(\mathbf{x}) \), where \( \mathbf{x} \in \mathbb{R}^n \) represents a point in n-dimensional space. Mathematically, \(f_{Noise}(\mathbf{x})\) is defined as a random field with a normal distribution:\[ f_{Noise}(\mathbf{x}) \sim \mathcal{N}(0, \sigma^2) \] where \( \sigma \) is the variance of the distribution. In our case, \( \sigma = 100 \). The dataset spans a specified input range in each dimension of \( [0, 1] \). Furthermore, the sample rate in each dimension is set to $20$ for the following experiments. To modulate the complexity of the function, a low-pass filter is applied. This is achieved by limiting the frequency range of the white noise. Let \[F_{Noise}(\mathbf{f}) = \mathcal{F}(f_{Noise}(\mathbf{x}) ) \] represent the Fourier transform of the white noise, converting it to the frequency domain. A cut-off frequency, \( f_c \), is defined to limit the frequencies. The filtered signal in the frequency domain, \( F_{Noise,f_{c}}(\mathbf{f}) \), is given by:
 \[ F_{Noise,f_{c}}(\mathbf{f}) = \begin{cases} 
  F(\mathbf{f}) & \text{if } |\mathbf{f}| \leq f_c \\
  0 & \text{otherwise} \end{cases} \] where \(\mathbf{f}\) represents the frequency vector in the frequency domain. The filtered signal in the frequency domain is then converted back to the spatial domain using the inverse Fourier transform:
 \[ f_{Noise,f_{c}}(\mathbf{x}) = \mathcal{F}^{-1}(F_{Noise,f_{c}}(\mathbf{f})) \]
 where \( f_{Noise,f_{c}}(\mathbf{x}) \) represents the low-pass filtered white noise in the spatial domain. The usage of the isotropic filter ensures the nonexistence of any axial bias in the function.

In traditional discrete signal processing, the frequency of a signal is inherently tied to the sample rate, with the highest representable frequency being half the sampling rate, known as the Nyquist frequency. However, in the case of our white noise regression model, the frequency definition deviates from this norm. Here, the frequency is not determined by the sampling rate but is instead dependent on a unit step along each axis of the n-dimensional space. This approach allows us to define a more comprehensible frequency representation. Nevertheless, the upper cut-off frequency limit is still tied to the Nyquist frequency to avoid aliasing effects. In our case, the Nyquist frequency is $f_N = f_a/2 = 10Hz$. An example of such a target function is depicted in Fig.~\ref{fig:white_noise} where the cut-off frequency is set to $4 Hz$.

\section{\ac{ffn} Hyperparameter settings}
\label{app:hyperparameter_settings}
\begin{figure*}[b]  
  \begin{subfigure}{0.58\columnwidth}
    \includegraphics[width=0.75\columnwidth]{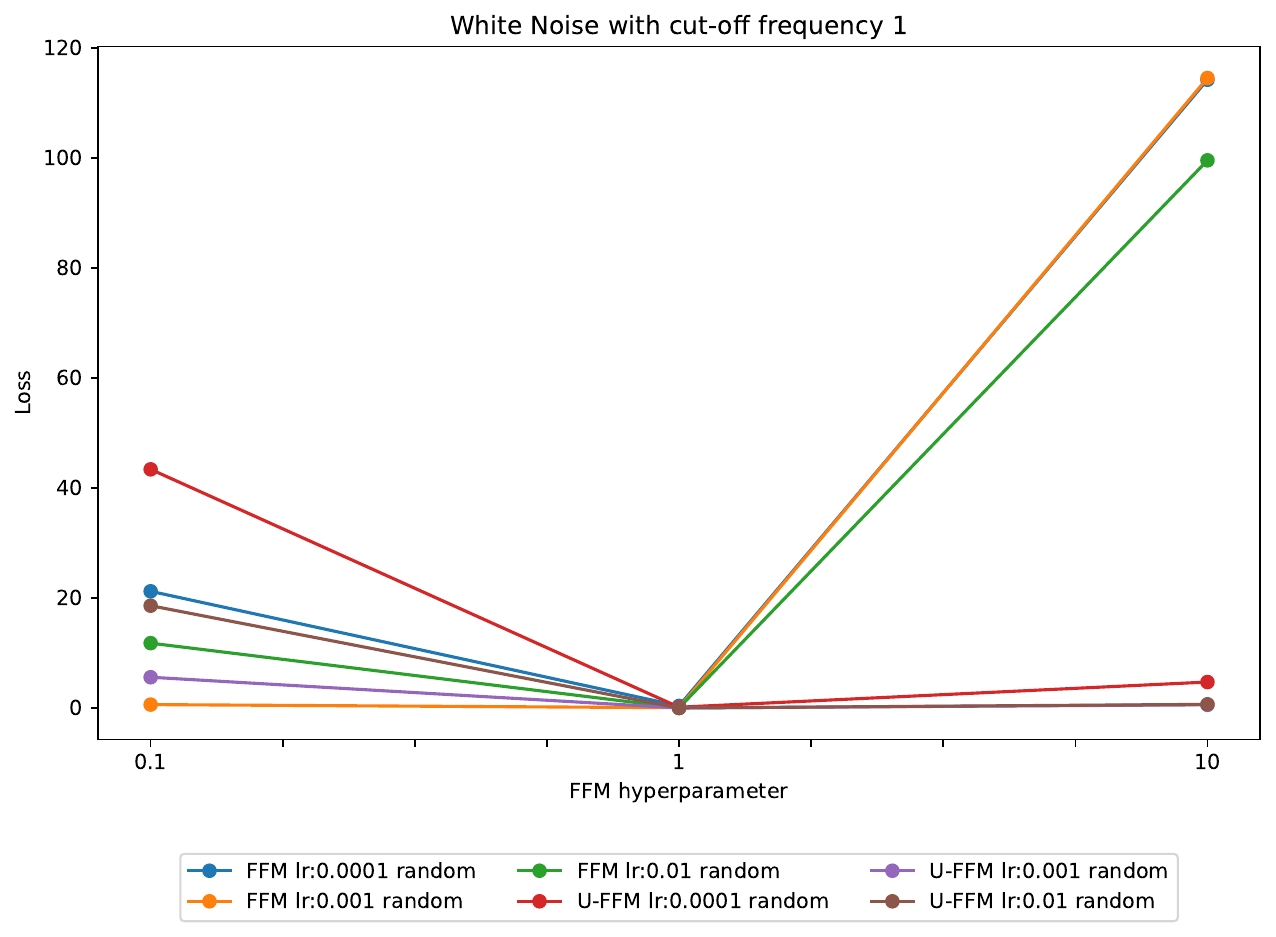}
    \label{fig:frequency_sweep_hyperparameters_1}
  \end{subfigure}
  \hfill
  \begin{subfigure}{0.58\columnwidth}
    \includegraphics[width=0.75\columnwidth]{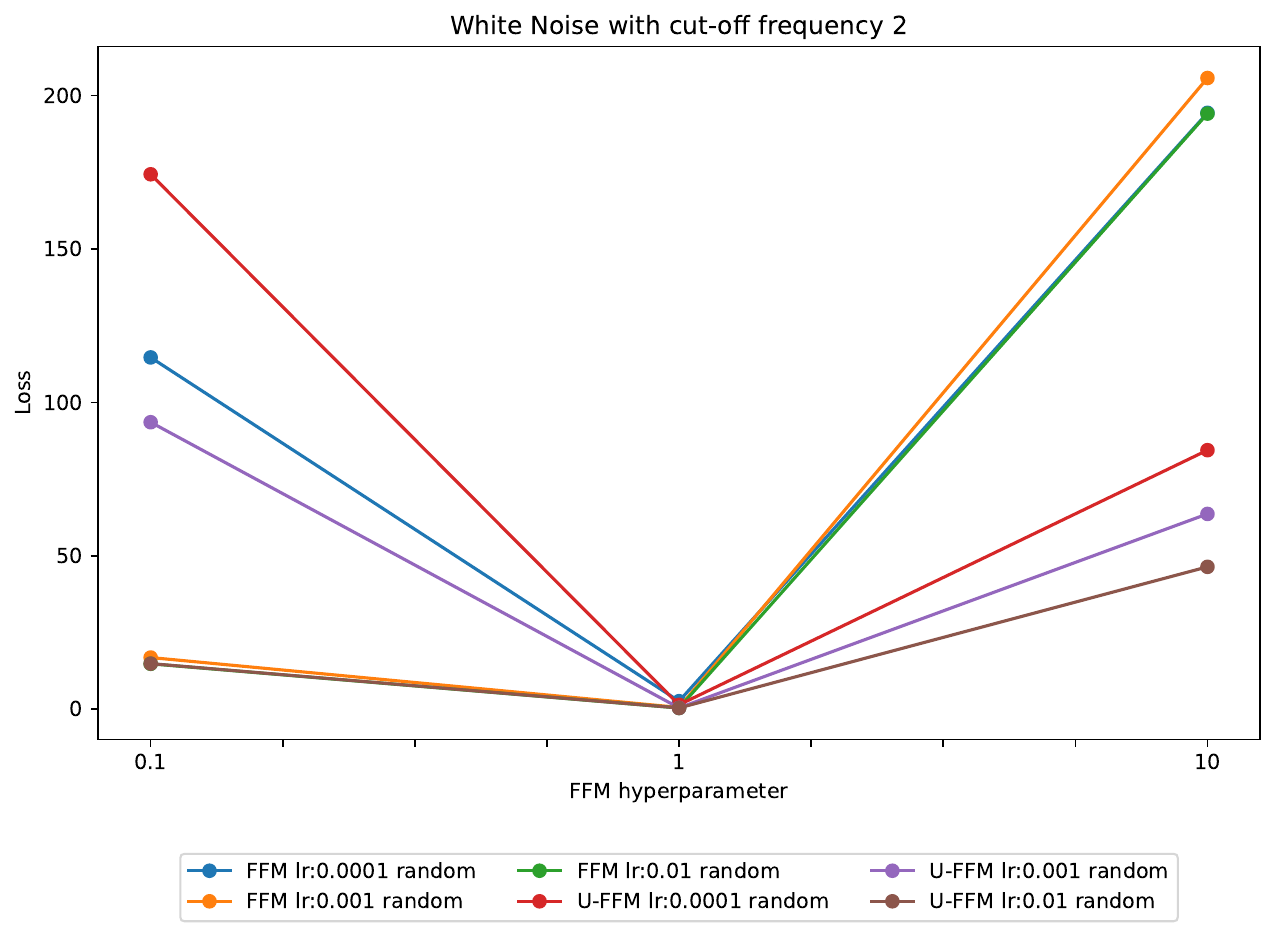}
    \label{fig:frequency_sweep_hyperparameters_2}
  \end{subfigure}
  \hfill
  \begin{subfigure}{0.58\columnwidth}
    \includegraphics[width=0.75\columnwidth]{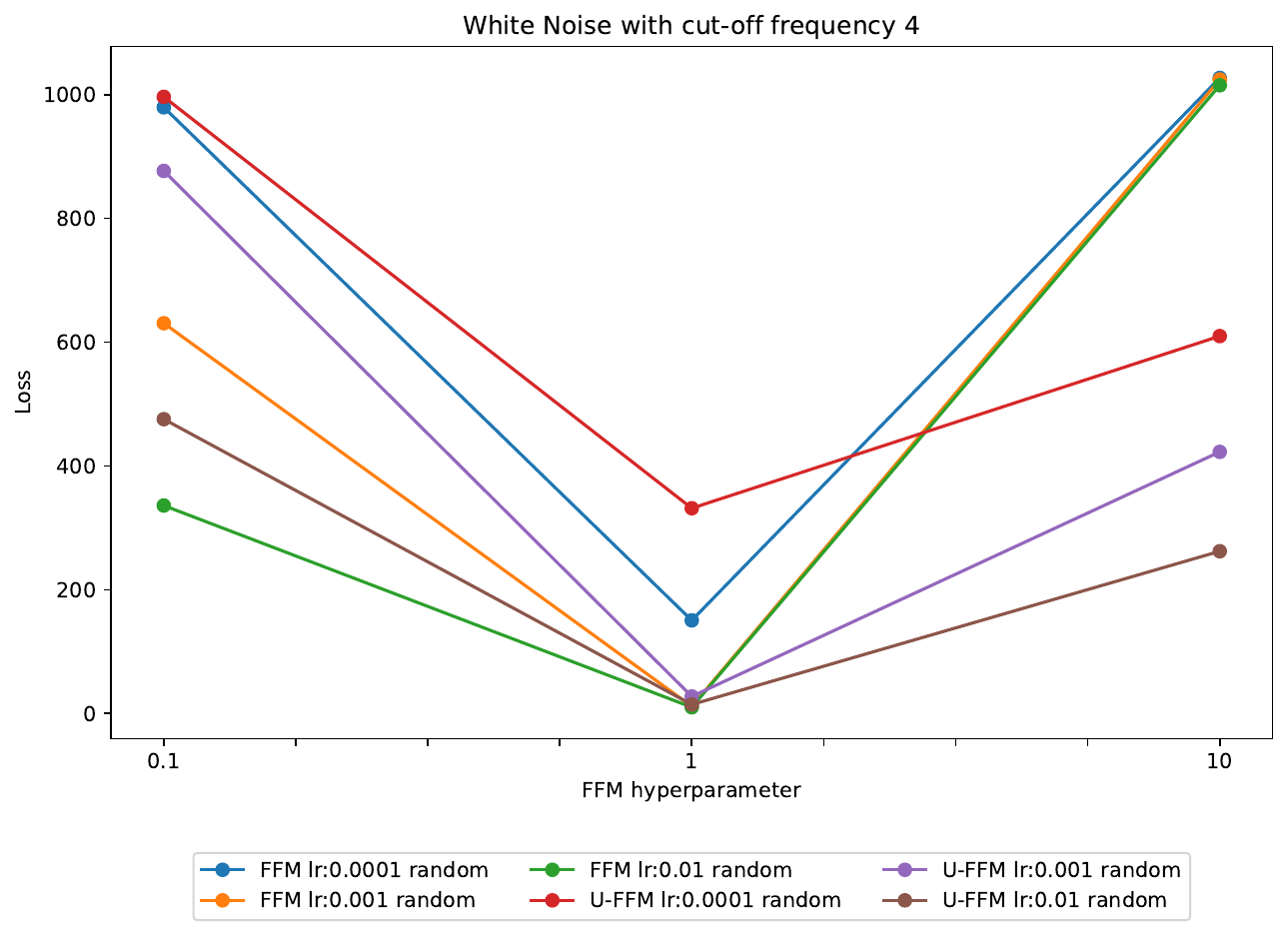}
    \label{fig:frequency_sweep_hyperparameters_4}
  \end{subfigure}%
  \hfill
  \begin{subfigure}{0.58\columnwidth}
    \includegraphics[width=0.75\columnwidth]{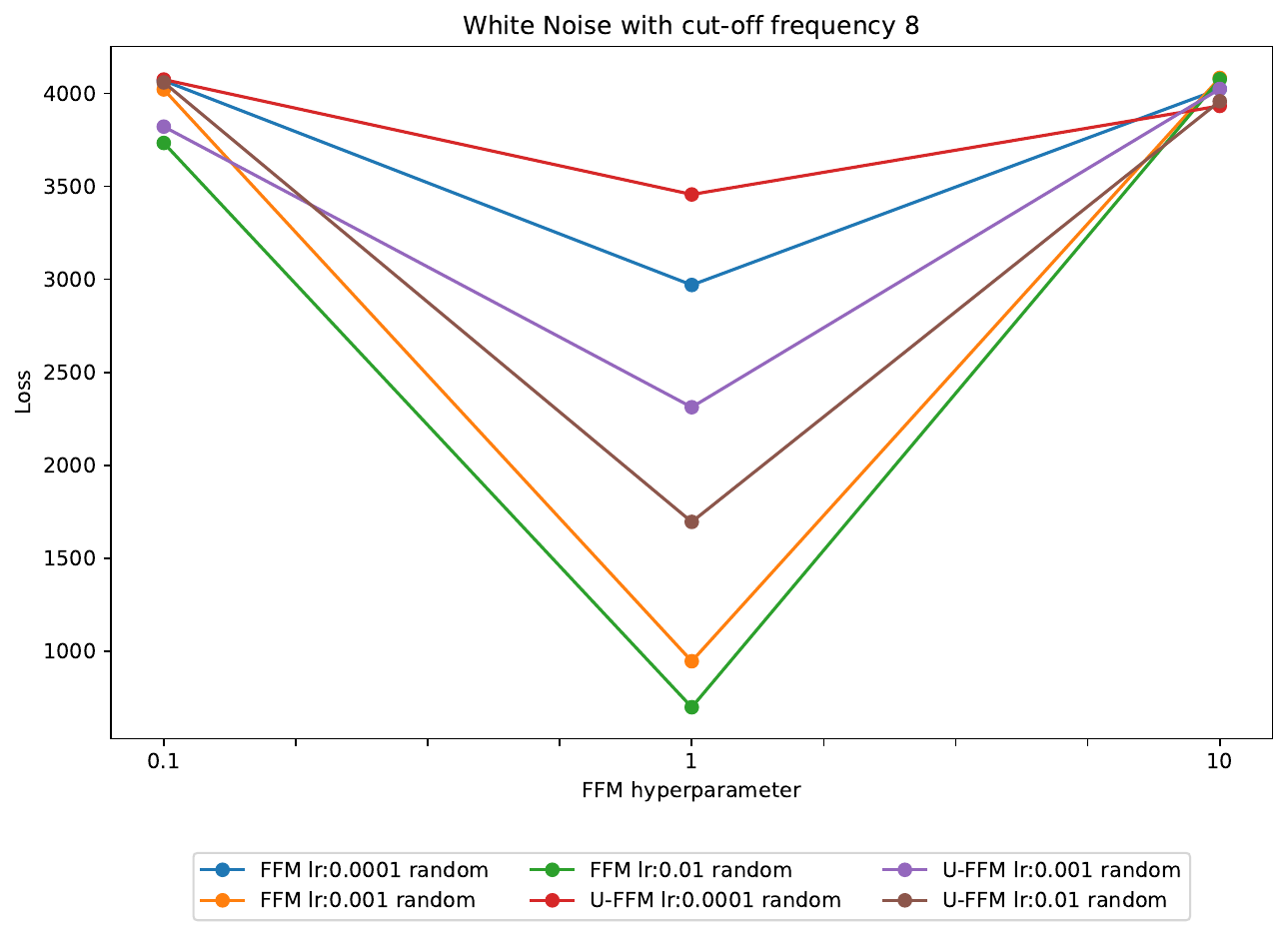}
    \label{fig:frequency_sweep_hyperparameters_8}
  \end{subfigure}
  \hfill
  \begin{subfigure}{0.58\columnwidth}
    \includegraphics[width=0.75\columnwidth]{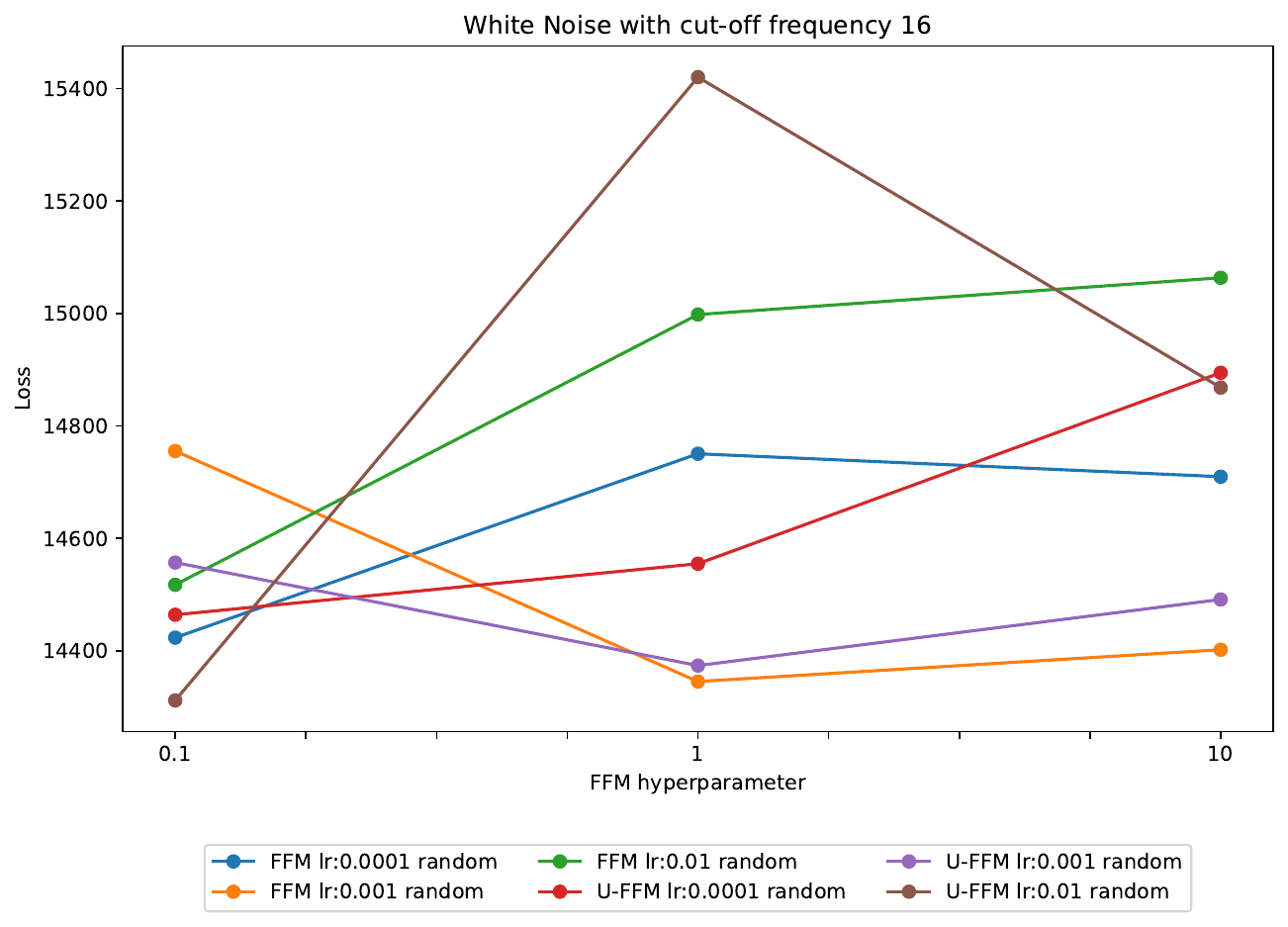}
    \label{fig:frequency_sweep_hyperparameters_16}
  \end{subfigure}
  \caption{Loss across different hyperparameter settings for white noise cut-off frequencies.}
  \label{fig:frequency_sweep_hyperparameters}
\end{figure*}

\begin{figure*}[b]  
  \begin{subfigure}[]{\columnwidth}
    \includegraphics[width=0.5\columnwidth]{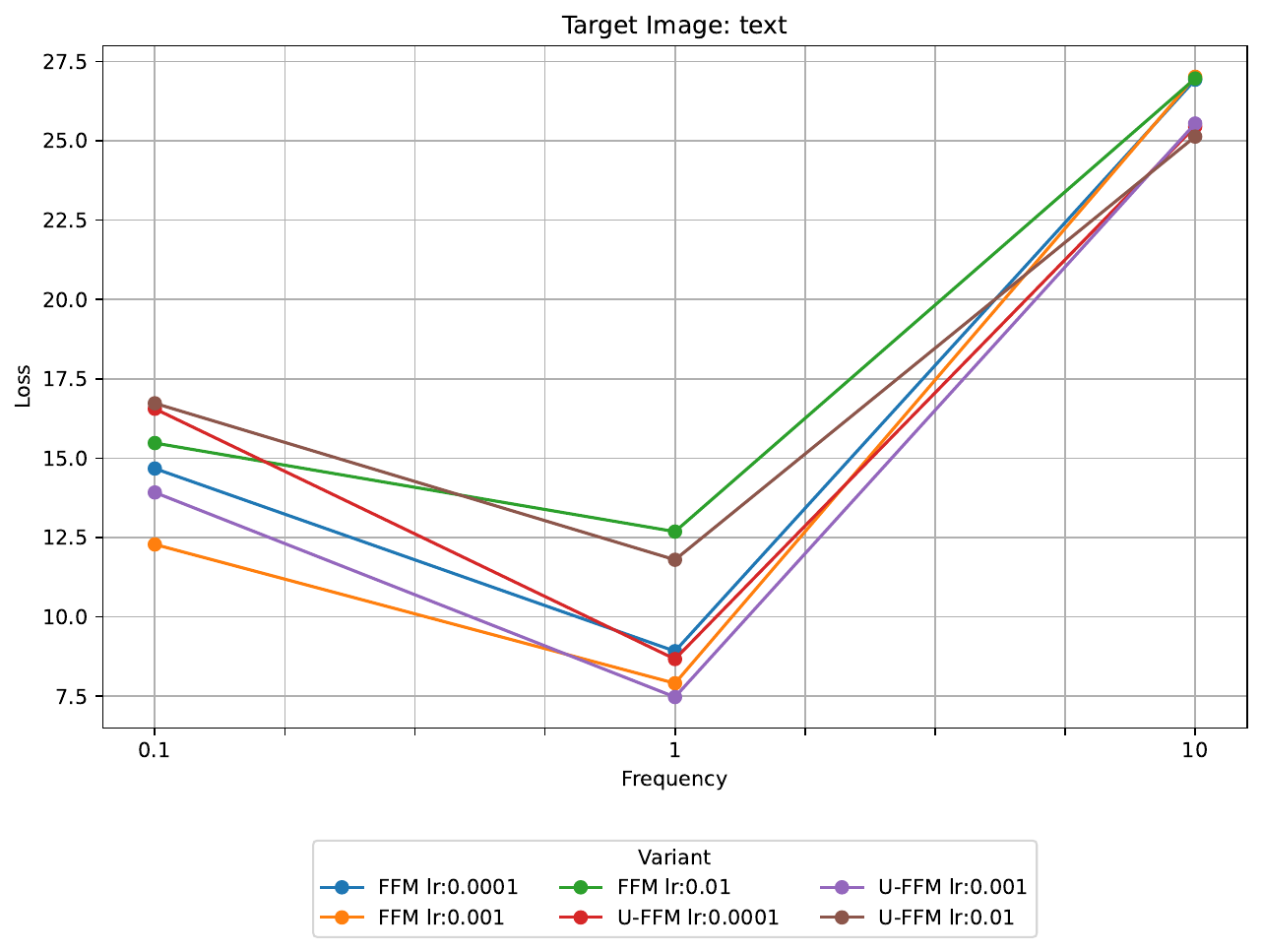}
    \label{fig:image_regression_hyperparameter_text}
  \end{subfigure}
  \begin{subfigure}[]{\columnwidth}
    \includegraphics[width=0.5\columnwidth]{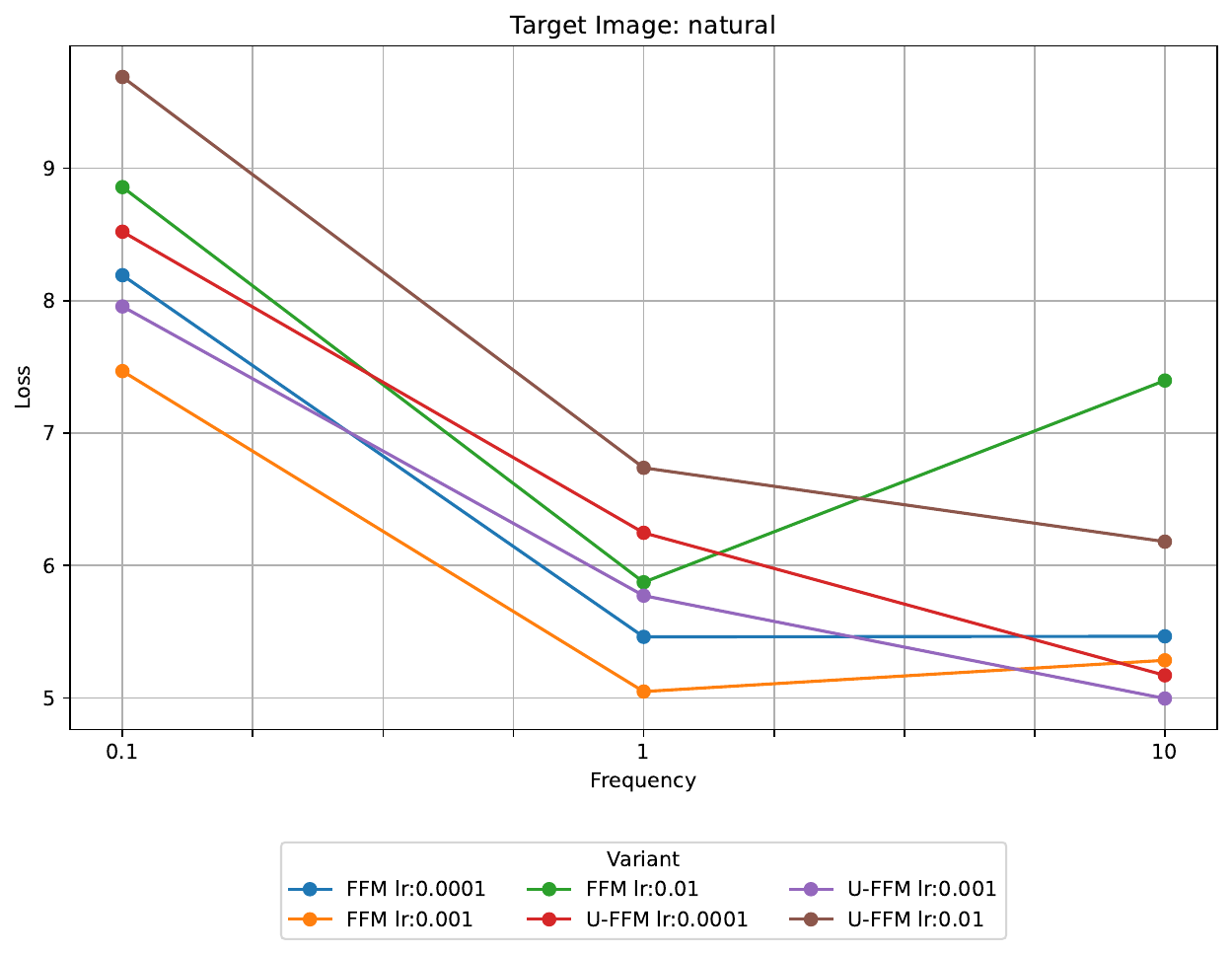}
    \label{fig:image_regression_hyperparameter_natural}
  \end{subfigure}

  \caption{Loss across different hyperparameter settings for different target images.}
  \label{fig:target_image_hyperparameters}
\end{figure*}

\begin{figure}[b]
    \includegraphics[width=0.5\columnwidth]{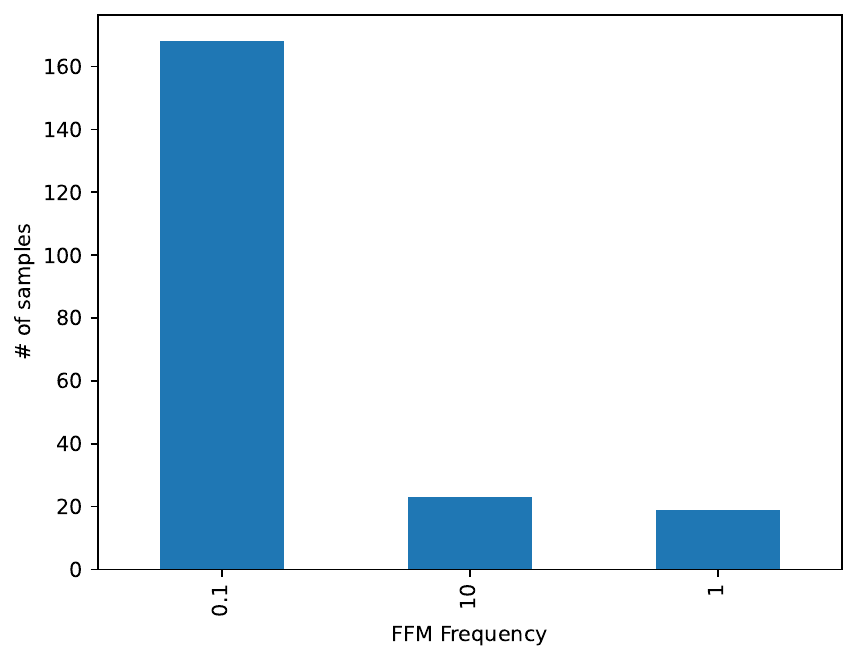}
    \caption{Loss across different hyperparameter settings for real-world datasets.\label{fig:dataset_ffm_hyperparameter}}
\end{figure}

For different Noise Cut-off frequencies, the optimal hyperparameter setting changes. Furthermore, it can be observed that the optimal hyperparameter setting depends on the data as well as other factors such as the learning rate and variant of the approach. In Fig.~\ref{fig:frequency_sweep_hyperparameters} is the performance per hyperparameter setting for different variants depicted. It can be observed that the setting of $1$ is not always optimal. Fig.~\ref{fig:target_image_hyperparameters} shows that for different target images in an image regression task, the optimal hyperparameter setting might change as well. In Fig.~\ref{fig:dataset_ffm_hyperparameter} the occurrence frequency of an optimal hyperparameter setting is depicted. Suggesting that lower values are more often the best setting among the evaluated settings. This also suggests that not always the optimal hyperparameter has been found.




\section{Dataset Regression}
\label{app:dataset_regression}
The detailed results of the dataset regression can be found in the following table \ref{tab:full_dataset_avr}.

\begin{table}
\small

\caption{Scaled Loss per Dataset.\label{tab:full_dataset_avr}}
\begin{tabular}{lrrrrrrrrr}
\toprule
 & SVR & GradBoost & MLP & M-RBF & M-RBF fixed & U-RBF & U-RBF fixed & FFN & U-FFN \\
\midrule
556\_analcatdata\_apnea2 & 8.52 & \textbf{1.00} & 2.29 & 8.07 & 7.83 & 1.15 & 3.53 & 7.95 & 1.13 \\
631\_fri\_c1\_500\_5 & 3.11 & 2.17 & 1.01 & 1.54 & 19.91 & 1.32 & 1.29 & 1.75 & \textbf{1.00} \\
601\_fri\_c1\_250\_5 & 4.52 & 2.07 & 1.24 & 1.93 & 18.13 & 1.08 & 1.02 & 2.57 & \textbf{1.00} \\
519\_vinnie & 1.65 & 1.12 & 1.01 & nan & 4.18 & 1.00 & \textbf{1.00} & 1.22 & 1.12 \\
485\_analcatdata\_vehicle & 2.65 & \textbf{1.00} & 1.25 & 1.40 & 2.46 & 1.33 & 1.31 & 1.27 & 1.11 \\
613\_fri\_c3\_250\_5 & 3.31 & 2.43 & 1.11 & 1.66 & 16.45 & 1.39 & 1.22 & 1.66 & \textbf{1.00} \\
1030\_ERA & \textbf{1.00} & 1.04 & 1.96 & 2.06 & 3.20 & 2.01 & 2.01 & 2.08 & 2.04 \\
597\_fri\_c2\_500\_5 & 3.57 & 2.20 & 1.10 & 1.16 & 24.70 & 1.22 & 1.21 & 1.60 & \textbf{1.00} \\
712\_chscase\_geyser1 & 4.19 & 1.37 & 1.29 & 1.99 & 4.48 & \textbf{1.00} & 1.01 & 1.33 & 1.16 \\
1027\_ESL & 1.05 & 1.09 & \textbf{1.00} & 1.09 & 6.59 & 1.09 & 1.11 & 1.15 & 1.01 \\
523\_analcatdata\_neavote & 3.36 & 2.74 & 1.30 & 1.32 & 15.38 & 1.41 & 1.93 & 1.99 & \textbf{1.00} \\
529\_pollen & \textbf{1.00} & 1.03 & 5.87 & nan & 29.27 & 6.23 & 6.27 & 30.10 & 7.63 \\
599\_fri\_c2\_1000\_5 & 1.81 & 1.17 & 1.01 & 1.21 & 36.65 & 1.09 & 1.09 & 1.43 & \textbf{1.00} \\
596\_fri\_c2\_250\_5 & 6.02 & 2.70 & 1.20 & 1.93 & 19.35 & 1.26 & 1.39 & 2.10 & \textbf{1.00} \\
557\_analcatdata\_apnea1 & 12.13 & 1.07 & 2.18 & 11.38 & 11.19 & \textbf{1.00} & 4.73 & 11.26 & 1.25 \\
624\_fri\_c0\_100\_5 & 1.72 & 1.75 & 1.21 & 1.99 & 8.26 & 1.22 & 1.26 & 2.49 & \textbf{1.00} \\
663\_rabe\_266 & 833.16 & 5.79 & 3.41 & nan & 1179.05 & \textbf{1.00} & 1.07 & 1569.11 & 14.96 \\
banana & 1.00 & \textbf{1.00} & 7.53 & 7.80 & 19.12 & 7.54 & 7.52 & 7.46 & 7.68 \\
228\_elusage & 4.68 & 1.89 & 1.48 & nan & 6.22 & 1.05 & \textbf{1.00} & 5.55 & 2.86 \\
579\_fri\_c0\_250\_5 & 1.46 & 2.11 & 1.04 & 1.29 & 10.77 & 1.31 & 1.41 & 2.11 & \textbf{1.00} \\
678\_visualizing\_environmental & \textbf{1.00} & 1.07 & 1.33 & nan & 1.35 & 1.05 & 1.06 & 1.35 & 1.25 \\
1096\_FacultySalaries & 5.55 & 1.91 & \textbf{1.00} & nan & 9.86 & 1.42 & 1.46 & 12.98 & 7.46 \\
1029\_LEV & \textbf{1.00} & 1.05 & 2.01 & 2.01 & 5.13 & 2.02 & 2.02 & 2.15 & 2.01 \\
210\_cloud & 1.34 & 1.02 & \textbf{1.00} & 1.26 & 3.99 & 1.20 & 1.29 & 1.69 & 1.16 \\
594\_fri\_c2\_100\_5 & 2.35 & 1.87 & 1.42 & 1.70 & 5.97 & 1.14 & 1.19 & 2.62 & \textbf{1.00} \\
192\_vineyard & 1.45 & 1.68 & 1.09 & 1.01 & 3.51 & \textbf{1.00} & 1.10 & 1.31 & 1.29 \\
609\_fri\_c0\_1000\_5 & \textbf{1.00} & 1.56 & 1.98 & 1.92 & 33.51 & 1.69 & 1.74 & 2.54 & 1.78 \\
611\_fri\_c3\_100\_5 & 3.42 & 4.77 & 1.34 & 3.23 & 15.22 & 1.55 & 1.49 & 2.44 & \textbf{1.00} \\
649\_fri\_c0\_500\_5 & 1.25 & 1.87 & 1.04 & 1.01 & 15.10 & 1.16 & 1.15 & 1.44 & \textbf{1.00} \\
617\_fri\_c3\_500\_5 & 4.36 & 2.47 & 1.13 & 1.49 & 22.76 & 1.03 & \textbf{1.00} & 1.62 & 1.22 \\
656\_fri\_c1\_100\_5 & 2.32 & 2.14 & 2.51 & 1.57 & 10.24 & 1.54 & 1.50 & 3.23 & \textbf{1.00} \\
612\_fri\_c1\_1000\_5 & 1.91 & 1.26 & \textbf{1.00} & 1.34 & 34.40 & 1.13 & 1.12 & 1.54 & 1.01 \\
628\_fri\_c3\_1000\_5 & 1.96 & 1.42 & 1.22 & 1.47 & 34.19 & \textbf{1.00} & 1.02 & 1.36 & 1.17 \\
690\_visualizing\_galaxy & 24.14 & \textbf{1.00} & 4.04 & nan & 141.64 & 1.34 & 1.42 & 88.25 & 8.24 \\
687\_sleuth\_ex1605 & 2.29 & \textbf{1.00} & 1.18 & nan & 3.06 & 1.09 & 1.10 & 4.05 & 2.25 \\
\bottomrule
\end{tabular}
\end{table}

\end{document}